\documentclass[11pt]{article}
\usepackage[margin=1in]{geometry}
\usepackage{amsmath,amsfonts,amssymb}
\usepackage{graphicx}
\usepackage{hyperref}
\usepackage{subcaption}
\usepackage{wrapfig}
\usepackage{amsmath,amssymb}
\usepackage{booktabs}
\usepackage{adjustbox}
\usepackage{makecell}
\usepackage{multirow}
\usepackage{caption}
\usepackage{xcolor}
\usepackage[table]{xcolor}

\newcommand{\instA}{\thanks{Institute of Robotics and Machine Intelligence, Poznan University of Technology}}

\title{Pointy – A Lightweight Transformer for Point Cloud Foundation Models}
\author{Konrad Szafer\instA ~ \href{https://orcid.org/0009-0005-6900-0484}{ORCID: 0009-0005-6900-0484} \and
Marek Kraft\footnotemark[1] ~ \href{https://orcid.org/0000-0001-6483-2357}{ORCID: 0000-0001-6483-2357} \and
Dominik Belter\footnotemark[1] ~ \href{https://orcid.org/0000-0003-3002-9747}{ORCID: 0000-0003-3002-9747}}
\date{}

\begin{document}
\maketitle

\begin{abstract}
Foundation models for point cloud data have recently grown in capability, often leveraging extensive representation learning from language or vision. In this work, we take a more controlled approach by introducing a lightweight transformer-based point cloud architecture. In contrast to the heavy reliance on cross-modal supervision, our model is trained only on 39k point clouds - yet it outperforms several larger foundation models trained on over 200k training samples. Interestingly, our method approaches state-of-the-art results from models that have seen over a million point clouds, images, and text samples, demonstrating the value of a carefully curated training setup and architecture. To ensure rigorous evaluation, we conduct a comprehensive replication study that standardizes the training regime and benchmarks across multiple point cloud architectures. This unified experimental framework isolates the impact of architectural choices, allowing for transparent comparisons and highlighting the benefits of our design and other tokenizer-free architectures. Our results show that simple backbones can deliver competitive results to more complex or data-rich strategies\footnote{The implementation, including code, pre-trained models, and training protocols, is available at \url{https://github.com/KonradSzafer/Pointy}.}.
\end{abstract}

\smallskip

\bigskip
\small\noindent\textit{To appear in the proceedings of ACIVS 2025. 
An earlier version was presented at the SCI-FM workshop at ICLR 2025.}

\section{Introduction}
\label{sec:introduction}

3D representation learning has grown increasingly important across domains such as robotics, augmented reality, and computational design. Recent advances in model architectures \cite{guo2021pct,zhao2021point,engel2021point,pang2022masked,yu2022point}, representation learning \cite{xue2024ulip,liu2024openshape,maoopendlign,Zhang_2024_CVPR,zhou2023uni3d}, and even multimodal large language models that incorporate point clouds \cite{guo2023point,xu2024pointllm,liu2024uni3d} have pushed the boundaries in 3D perception. However, robust 3D feature extraction remains challenging. Additionally, current training protocols and evaluation environments for point cloud architectures diverge in the literature, hindering fair and transparent method comparison. Consequently, many architectural choices are guided by experimental setups differing in data regimes, preprocessing, or hyperparameter tuning, limiting unbiased conclusions about any given design's strengths.

Meanwhile, the concept of foundation models has taken hold in other domains such as natural language processing \cite{devlin2018bert,radford2019language,brown2020language}, computer vision \cite{he2016deep,DosovitskiyB0WZ21}, and time series \cite{goswami2024moment}.

For instance, in computer vision, architectural innovations such as the Swin Transformer~\cite{liu2021swin} have propelled state-of-the-art performance; however, as noted by Oquab et al. \cite{oquab2024dinov2}, it remains challenging to disentangle whether these gains primarily stem from large-scale datasets, advanced architectures, or refined training strategies. The landscape for point cloud data is also evolving rapidly, with numerous approaches now incorporating cross-modal learning of images or text, often using datasets of hundreds of thousands to millions of samples. While these efforts have unquestionably expanded the frontiers of 3D representation learning, they also highlight the urgent need for standardized, controlled experiments to isolate the underlying drivers of model performance.

In this work, we propose a lightweight, transformer-based backbone for point cloud processing that operates directly on point coordinates without requiring a separately trained tokenizer. Despite using just 39k shapes and a simple classification objective, our method outperforms larger foundation models trained on over 200k point clouds. More surprisingly, it approaches state-of-the-art results achieved by models trained on over a million samples from point cloud, image, and text representations. To ensure a robust and fair evaluation, we conduct a replication study under a unified training and benchmarking regime. We compare our approach with other tokenizer-free architectures while applying the same preprocessing steps, optimizer settings, and hyperparameter schedules. 

\textbf{Our contributions are threefold:}
\begin{itemize}
    \item 1. \textbf{Comprehensive replication study:} We propose a benchmarking system for point cloud foundation models to evaluate popular transformer backbones under identical training regimes and evaluation metrics, providing an apples-to-apples comparison of their strengths and limitations.
    
    \item 2. \textbf{Lightweight yet competitive model:} We present a simple transformer-based backbone that excels without relying on massive cross-modal supervision or large data sets.
    
    \item 3. \textbf{Controlled pre-training:} We show that limited data can compete with much larger data-driven approaches, providing practical insights into scaling strategies and design trade-offs.

\end{itemize}

\section{Related Work}
\label{sec:related}

\textbf{Point Cloud Deep Learning Architectures.} Early work like PointNet and PointNet++ \cite{qi2017pointnet,qi2017pointnet++}, laid the foundation for deep learning on irregular 3D data by introducing permutation invariant architectures that operate directly on unordered point sets. Building on these seminal ideas, subsequent methods have explored graph-based models \cite{wang2019dynamic} and convolutional operators to better capture local and global geometric features. More recently, transformer-based approaches \cite{guo2021pct,zhao2021point,engel2021point,pang2022masked} have emerged as a promising direction, exploiting self-attention to model long-range dependencies inherent in point clouds. In parallel, advances in self-supervised representation learning \cite{pang2022masked,qi2023contrast,yu2022point} have further refined these architectures and training paradigms by enabling robust feature extraction even under limited supervision.

\textbf{Foundation Models and Cross-Modal Representation Learning.} 
The emergence of foundation models in natural language processing \cite{devlin2018bert,radford2019language,brown2020language}, computer vision \cite{he2016deep,DosovitskiyB0WZ21}, and other modalities \cite{goswami2024moment} has underscored the transformative impact of large-scale pre-training and unified architectures. Inspired by these successes, recent efforts in 3D representation learning have begun to incorporate cross-modal supervision, aligning point clouds with image and text modalities \cite{guo2023point,xu2024pointllm,liu2024uni3d} to improve performance. 

However, the significant data requirements and complex training pipelines of these multimodal models raise questions about their true success drivers.

\begin{table}[t!]
\caption{\textbf{Comparison of classification performance across ModelNet40 and ScanObjectNN benchmarks}. We report overall accuracy (OA), model parameters, and FLOPs (in G), with all models trained under identical conditions: batch size 16 for 100 epochs, points normalized to the [-1,1] range, and random z-axis rotation. Results demonstrate the competitive performance of our lightweight architecture (3.0M parameters, 16.2G FLOPs) across diverse point cloud datasets. Models marked with \textbf{[T]} denote transformer-based architectures, while those with \textbf{[ST]} incorporate an unchanged transformer block.}

\centering
\resizebox{\linewidth}{!}{%
\begin{tabular}{l c c c c c}
\toprule
\multirow{2}{*}{Model} & \multirow{2}{*}{\#Points} & \multirow{2}{*}{Param. (M) $\downarrow$} & \multirow{2}{*}{FLOPs (G) $\downarrow$} & ModelNet40 & ScanObjectNN \\
& & & & OA (\%) $\uparrow$ & OA (\%) $\uparrow$ \\
\midrule
PointNet \cite{qi2017pointnet} & 2k & 1.7 & 8.1 & \textbf{91.6} & \underline{80.3} \\
PointNet++ \cite{qi2017pointnet++} & 2k & 1.7 & 8.1 & \underline{90.7} & \textbf{81.7} \\
DGCNN \cite{wang2019dynamic} & 2k & 1.4 & 17.8 & 90.8 & 71.5 \\
PointMLP \cite{ma2022rethinking} & 2k & 13.1 & 62.9 & \textbf{91.6} & 78.8 \\
\midrule
\multicolumn{6}{c}{Transformer-based} \\
\midrule
\textbf{[T]} PCT \cite{guo2021pct} & 2k & 2.9 & 4.3 & 90.1 & 73.0 \\
\textbf{[T]} PointTransformer \cite{zhao2021point} & 2k & 9.5 & 73.6 & 90.1 & 76.3 \\
\textbf{[T]} PointTransformer \cite{engel2021point} & 2k & 22.0 & 15.2 & 88.0 & 64.6 \\
\textbf{[ST]} PointMAE \cite{pang2022masked} & 2k & 21.8 & 4.7 & 89.8 & 78.0 \\  
\midrule
\textbf{[ST]} Our - Small & 2k & 3.0 & 16.2 & \underline{90.4} & \textbf{80.0} \\
\textbf{[ST]} Our - Base & 2k & 19.4 & 18.0 & \textbf{90.6} & \underline{78.5} \\
\bottomrule
\end{tabular}
}
\label{tab:classification_table}
\end{table}

\begin{figure}[ht]
  \centering
  \includegraphics[width=\linewidth]{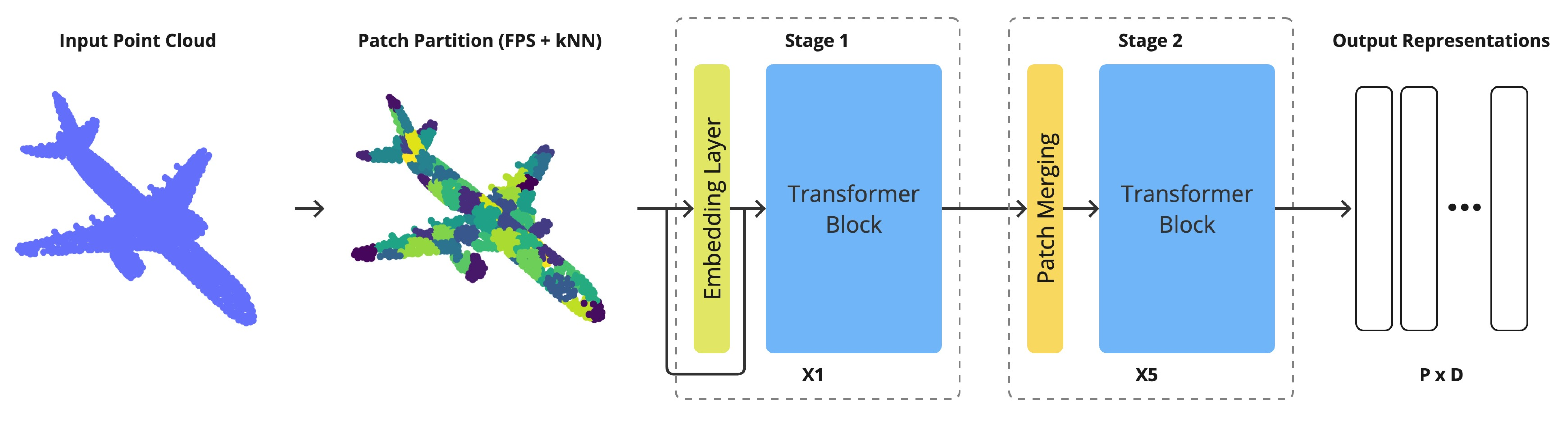}
  \caption{Architecture Pointy -- transformer backbone for point cloud processing. The model takes raw point cloud data as input, applies patch partitioning using Farthest Point Sampling (FPS), and k-Nearest Neighbors (kNN). Raw point features are preserved through residual connections alongside learned patch embeddings in the embedding layer based on PointNet. The architecture consists of token merging operations between adjacent tokens after each transformer block. This hierarchical design enables local and global feature learning through progressive patch merging. The final output produces P$\times$D dimensional representations, where P is the number of patches and D is the embedding dimension.}
  \label{fig:architecture}
\end{figure}

\section{Pointy}
\label{sec:pointy}

We present \textit{Pointy}, our lightweight transformer-based model for point cloud processing. The main goal is to jointly learn the patch-level embeddings and the global representation in a single architecture while keeping the core transformer block unchanged. Unlike some of the previous works that rely on replaced attention layers or separate tokenization schemes, our approach learns embeddings directly from points, complemented by positional information. An overview of the proposed pipeline is given in Fig.~\ref{fig:architecture}.

\subsection{Point Embeddings}
\label{sec:embedding}

To encode the raw point cloud into a compact set of feature tokens, we adopt a simplified PointNet-like \cite{qi2017pointnet} embedding module, scaled down to fit our goal of lightweight design. Inspired by residual connections \cite{he2016deep}, we include them, so absolute coordinates pass with learned features, preserving geometric information. In addition, we add a learnable positional embedding \cite{DosovitskiyB0WZ21}, which provides encoding of the spatial arrangement of the patches.

Following common practice in vision transformers, we aggregate local patches of points before feeding them into the backbone. Specifically, we use Farthest Point Sampling (FPS) for anchor points, then group their $k$-nearest neighbors (kNN) as in \cite{zhao2021point}. Each neighborhood is transformed into a feature vector via our embedding layer and combined with a learned positional token. This tokenizer-free strategy reduces the complexity of handling irregular input while preserving fine-grained local geometry. Our method can be trained on $\mathbb{R}^{3 \times N}$ (3D coordinates) and $\mathbb{R}^{6 \times N}$ (3D coordinates + normals/color).

\begin{table}[t!]
\caption{\textbf{Zero-shot Evaluation of Transformer-based Models Pre-trained on the Objaverse-LVIS Subset.} All models are pre-trained on an 85/15 split of LVIS data (85\% training, 15\% testing) drawn from a curated 39k-sample subset spanning 1,156 LVIS-annotated object categories. Training is performed for 30 epochs under identical conditions—batch size 16, 2k points per sample normalized to the [-1,1] range, and random rotations about the z-axis. The checkpoints obtained after this training are subsequently used for zero-shot evaluation, ensuring a consistent and fair comparison of transferability.}
\label{tab:objaverse_lvis_transformers}
\centering
\resizebox{\linewidth}{!}{%
\begin{tabular}{l ccc c}
\toprule
\multirow{2}{*}{Model} & \multirow{2}{*}{\#Points} & \multirow{2}{*}{Param. (M) $\downarrow$} & \multirow{2}{*}{FLOPs (G) $\downarrow$} & Objaverse-LVIS \\
& & & & OA (\%) $\uparrow$ \\
\midrule
PCT \cite{guo2021pct}                  & 2k & 2.9  & 4.3  & \cellcolor{blue!10}\underline{36.3} \\
PointTransformer \cite{zhao2021point}  & 2k & 9.5  & 73.6 & 34.1 \\
PointTransformer \cite{engel2021point} & 2k & 22.0 & 15.2 & 8.5 \\
PointMAE \cite{pang2022masked}         & 2k & 21.8 & 4.7  & 34.9 \\
\midrule
Our - Small                            & 2k & 3.0  & 16.2 & \cellcolor{blue!10}\textbf{36.4} \\
Our - Base                             & 2k & 19.4 & 18.0 & \cellcolor{blue!10}\underline{36.3} \\
\bottomrule
\end{tabular}
}
\label{tab:pretraining}
\end{table}
\vspace{-0.3cm}

\subsection{Transformer Backbone}
\label{sec:backbone}

We feed the resulting embeddings into a hierarchical transformer designed to handle up to 64 patch tokens but easily adjustable through configurable merging ratios in one of its layers. The transformer consists of six layers of multi-head self-attention, with patch merging operations to progressively reduce the spatial footprint of the token set. Each layer uses a relatively low embedding dimension to attentional heads ratio (approx. 3:1), introducing an inductive bias that aligns with the 3D coordinate structure of point clouds and promotes spatially coherent attention. By default, we use the GeLU activation function and standard layer normalization throughout. All network weights are initialized using the Kaiming method \cite{he2015delving}.

We introduce two Pointy size variants for different resource constraints and pre-training requirements. The smaller model, with a 192-dimensional embedding, includes approximately 3M parameters, while the larger version, with 20M parameters, scales the embedding dimension to 510. Both variants deliver competitive performance across a variety of classification tasks and resolutions (Tab.~\ref{tab:classification_table} and ~\ref{tab:pretraining}). In addition, our models achieve very competitive performance on zero-shot tasks \ref{tab:zero_shot} when pre-trained on 39k point clouds, compared to the state-of-the-art foundation models trained on significantly larger datasets.

\begin{table*}[t!]
\caption{\textbf{Zero-shot 3D Classification on ModelNet40 and ScanObjectNN.} For this evaluation, we discard the classification head and directly extract the final-layer features from the 30th epoch checkpoint of transformer models pre-trained on the Objaverse-LVIS split (39k point clouds). We report Top-1, Top-3, and Top-5 accuracies on both ModelNet40 and ScanObjectNN, comparing our reproduced transformer baselines (e.g., PCT, two distinct PointTransformer models, and PointMAE) along with our own small and base variants against third-party foundation models. Despite the modest dataset size relative to other large-scale pre-trained models, nearly all transformer-based architectures score impressively well. Our models achieve the best performance among them, with the PCT closely following, while outperforming several methods pre-trained on up to 200k point clouds. Only two foundation models, trained on over one million point clouds and augmented by training with representations from additional modalities, record higher scores. Blue shading indicates transformer-based backbones replicated under our experimental setting, while yellow shading denotes reported values for foundation models from the literature.}
\centering
\small
\setlength{\tabcolsep}{6pt}
\renewcommand{\arraystretch}{1.3}
\adjustbox{max width=\linewidth}{
\begin{tabular}{l c c c c ccc ccc}
\toprule
\multirow{2}{*}{Model} & \multirow{2}{*}{Pre-train Method} & \multirow{2}{*}{Pre-train Dataset} & \multirow{2}{*}{\#Point Clouds} & \multirow{2}{*}{\#Points} & \multicolumn{3}{c}{ModelNet40} & \multicolumn{3}{c}{ScanObjectNN} \\
& & & & & Top-1 & Top-3 & Top-5 & Top-1 & Top-3 & Top-5 \\
\midrule
\multicolumn{11}{c}{Our Reproduction} \\
\midrule
PCT \cite{guo2021pct} & Classification & Objaverse-LVIS & 39k & 2k & 83.5 & \cellcolor{blue!10}\underline{92.1} & 93.4 & \cellcolor{blue!10}\underline{55.2} & \cellcolor{blue!10}\underline{73.3} & \cellcolor{blue!10}\underline{81.6} \\
PointTransformer \cite{zhao2021point} & Classification & Objaverse-LVIS & 39k & 2k & 82.4 & 91.3 & 93.9 & 49.2 & 71.6 & 78.4 \\
PointTransformer \cite{engel2021point} & Classification & Objaverse-LVIS & 39k & 2k & 33.8 & 55.3 & 64.8 & 7.2 & 15.6 & 22.8 \\
PointMAE \cite{pang2022masked} & Classification & Objaverse-LVIS & 39k & 2k & 81.4 & 91.6 & 93.8 & 48.1 & 67.3 & 77.2 \\
\midrule
Our - Small & Classification & Objaverse-LVIS & 39k & 2k & \cellcolor{blue!10}\underline{83.9} & \cellcolor{blue!10}\textbf{92.2} & \cellcolor{blue!10}\underline{93.8} & 53.1 & 72.6 & 80.7 \\
Our - Base  & Classification & Objaverse-LVIS & 39k & 2k & \cellcolor{blue!10}\textbf{85.3} & \cellcolor{blue!10}\underline{92.1} & \cellcolor{blue!10}\textbf{94.1} & \cellcolor{blue!10}\textbf{58.3} & \cellcolor{blue!10}\textbf{77.8} & \cellcolor{blue!10}\textbf{84.4} \\
\midrule
\multicolumn{11}{c}{Reported in Literature} \\
\midrule
ReCon \cite{qi2023contrast} & ReCon \cite{qi2023contrast} & ShapeNet & 200k & --- & 49.5 & --- & 81.2 & 2.7 & --- & 7.9 \\
Point-BERT \cite{yu2022point} & OpenShape \cite{liu2024openshape} & ShapeNet & 200k & --- & 70.3 & --- & 91.3 & 10.8 & --- & 25.0 \\
Point-BERT \cite{yu2022point} & OpenShape \cite{liu2024openshape} & \makecell{Objaverse\\+ ShapeNet} & $>$1M & --- & 82.6 & 84.4 & 96.9 & 46.5 & 66.1 & 76.3 \\
Point-BERT \cite{yu2022point} & ULIP-2 \cite{xue2024ulip} & ShapeNet & 200k & --- & 75.2 & --- & 95.0 & 16.4 & --- & 34.3 \\
PointBERT \cite{yu2022point} & \makecell{OpenShape \cite{liu2024openshape}\\+ Dlign \cite{maoopendlign}} & Ensemble & $>$1M & --- & 85.4 & \cellcolor{yellow!30}\underline{96.6} & \cellcolor{yellow!30}\underline{98.2} & 51.1 & 77.4 & 88.2 \\
\makecell[l]{SparseConv\\ \cite{graham2017submanifold}} & \makecell{OpenShape \cite{liu2024openshape}\\+ Dlign \cite{maoopendlign}} & ShapeNet & 200k & --- & 74.9 & 89.5 & 94.1 & 56.3 & 75.2 & 85.4 \\
PointBERT \cite{yu2022point} & \makecell{TAMM \cite{Zhang_2024_CVPR}\\+ Dlign \cite{maoopendlign}} & ShapeNet & $>$1M & --- & 73.7 & 89.1 & 92.2 & 57.3 & 73.6 & 82.3 \\
PointBERT \cite{yu2022point} & \makecell{TAMM \cite{Zhang_2024_CVPR}\\+ Dlign \cite{maoopendlign}} & Ensemble & $>$1M & --- & \cellcolor{yellow!30}\underline{86.2} & \cellcolor{yellow!30}\underline{96.6} & 97.5 & \cellcolor{yellow!30}\underline{60.5} & \cellcolor{yellow!30}\underline{82.5} & \cellcolor{yellow!30}\underline{90.4} \\
Transformer \cite{vaswani2017attention} & Uni3D \cite{zhou2023uni3d} & Ensemble & $>$1M & --- & \cellcolor{yellow!30}\textbf{88.2} & \cellcolor{yellow!30}\textbf{98.4} & \cellcolor{yellow!30}\textbf{99.3} & \cellcolor{yellow!30}\textbf{65.3} & \cellcolor{yellow!30}\textbf{85.5} & \cellcolor{yellow!30}\textbf{92.7} \\  
\bottomrule
\end{tabular}
}
\label{tab:zero_shot}
\end{table*}

\section{Experimental Setup}
\label{sec:setup}

We design a unified training environment to ensure fair comparisons across point cloud architectures. Unless otherwise specified, all methods use the same data resolution, normalization, augmentation strategies, loss function, and optimizer configuration. Below, we summarize implementation details used in our study.

\subsection{Datasets and Tasks}
\label{sec:datasets}

\paragraph{ModelNet40 \cite{wu20153d}.} The dataset comprises 12,308 CAD models (40 object classes) with a predefined standard training/test split. We sample 2,048 points from each shape and train each method for 100 epochs using identical hyperparameters and preprocessing. As is common practice, the metric reported is overall classification accuracy on the test set.

\paragraph{ScanObjectNN \cite{uy2019revisiting}.} Contains 2,890 real-world 3D scans with 15 object categories, introducing substantially higher complexity due to noise, clutter, and occlusions. As with ModelNet40, we sample 2,048 points per scan and train for 100 epochs with the same configuration. We report the best test accuracy achieved by each method.

\paragraph{Objaverse-LVIS Subset \cite{deitke2023objaverse}.} To examine performance on a larger-scale dataset while maintaining a controlled pre-training regime, we utilize a 39k-sample subset from Objaverse spanning 1,156 LVIS-annotated object categories. We maintain a consistent 85\%/15\% train-test split across all experimental runs. Each model is trained for 30 epochs before evaluating zero-shot transfer to ModelNet40 and ScanObjectNN.

\subsection{Baselines}
\label{sec:baselines}

We evaluate a diverse set of architectures, spanning both classical point cloud networks and transformer-based models. All implementations are sourced from official repositories or widely validated reproductions, preserving their original architectural designs and hyperparameters (e.g., hidden dimensions, number of layers, dropout ratios).

\begin{figure}[t!]
  \centering
  \begin{minipage}[b]{0.32\linewidth}
    \includegraphics[width=\linewidth]{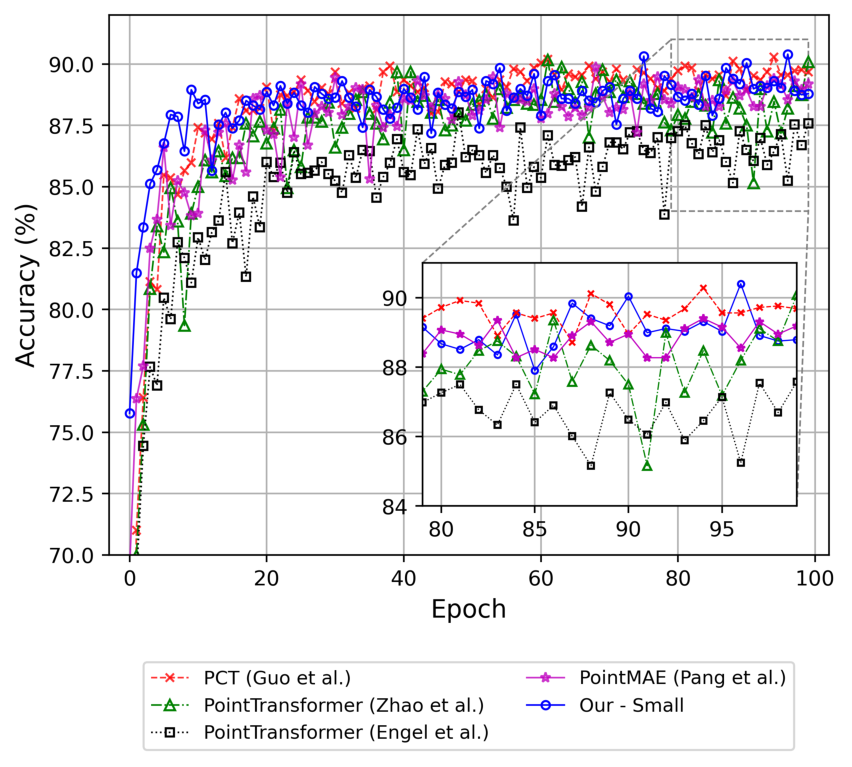}
    \subcaption{ModelNet40}
    \label{fig:modelnet40}
  \end{minipage}
  \hfill
  \begin{minipage}[b]{0.32\linewidth}
    \includegraphics[width=\linewidth]{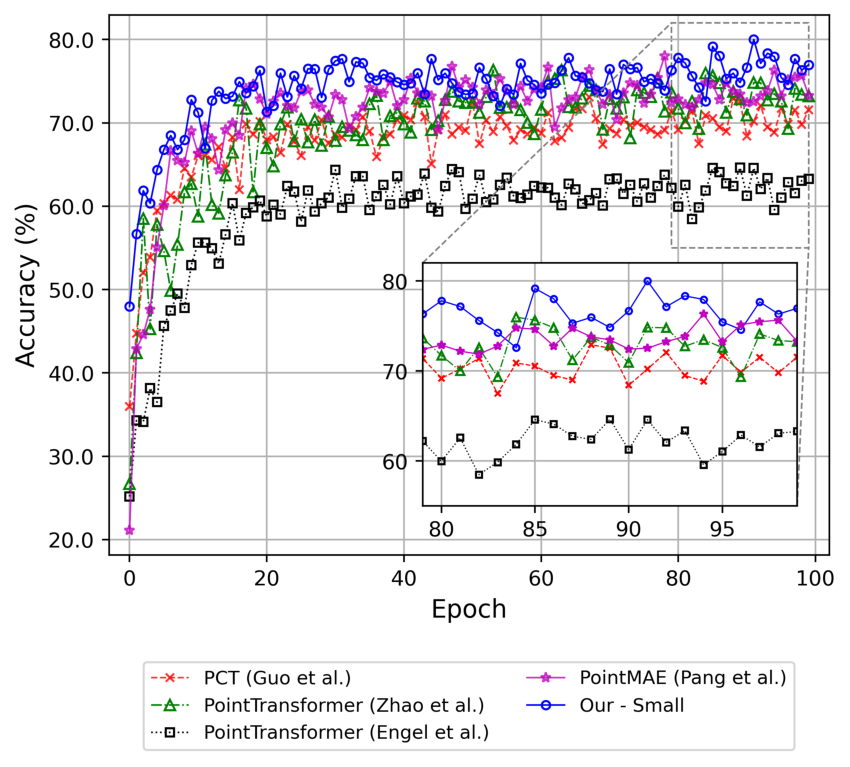}
    \subcaption{ScanObjectNN}
    \label{fig:scanobjectnn}
  \end{minipage}
  \hfill
  \begin{minipage}[b]{0.32\linewidth}
    \includegraphics[width=\linewidth]{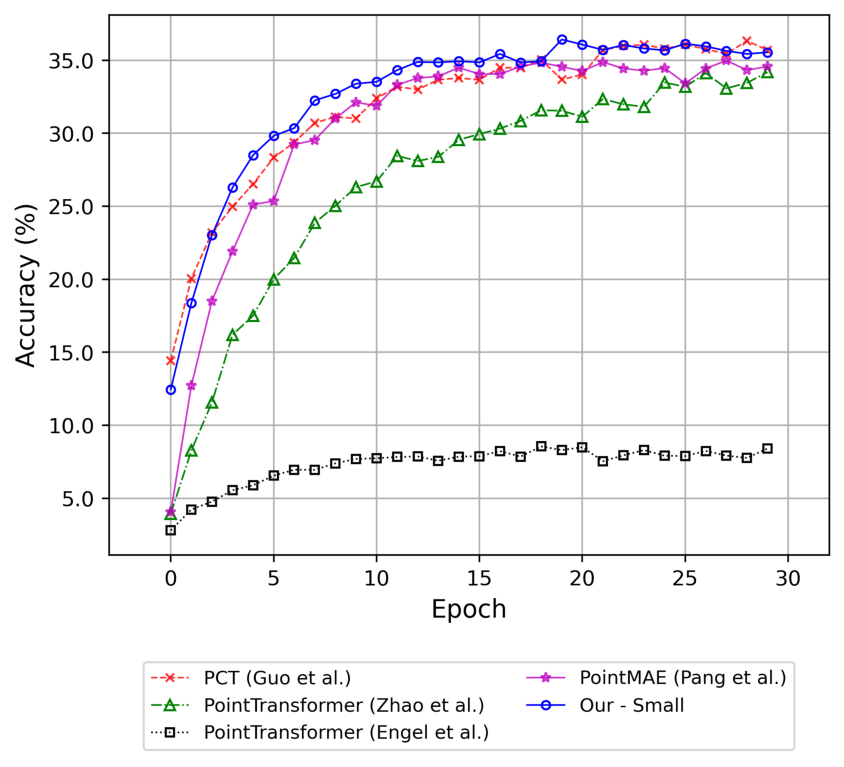}
    \subcaption{Objaverse-LVIS}
    \label{fig:objaverse}
  \end{minipage}
  \caption{\textbf{Training dynamics of different transformer-based models on ModelNet40, ScanObjectNN, and Objaverse-LVIS datasets.} The plots show overall classification accuracy (\%) versus training epochs for our proposed small model compared to existing approaches: PCT \cite{guo2021pct}, PointMAE \cite{pang2022masked}, and two variants of PointTransformer \cite{zhao2021point} and \cite{engel2021point}. Our method demonstrates faster convergence across all datasets while achieving competitive or superior final accuracy. The inset plots for (a) and (b) show detailed performance in later epochs. Notable is the consistent underperformance of PointTransformer~\cite{engel2021point} relative to other methods across all datasets.}
  \label{fig:training_plots}
\end{figure}

\subsection{Implementation Details}
\label{sec:implementation}

We maintain consistent experimental conditions across all evaluations: each shape is preprocessed by uniformly sampling 2,048 points and normalized to the [-1,1] range, with random z-axis rotation as the only augmentation and no voting-based inference. The training utilizes the AdamW optimizer with a fixed learning rate of $1\times10^{-4}$ without scheduling, using a batch size of 16 and standard cross-entropy loss. Models are trained for up to 100 epochs on ModelNet40 and ScanObjectNN, and 30 epochs on Objaverse-LVIS, reporting the best-observed test accuracy.

\subsection{Results}
\label{sec:results}

\subsubsection{Classification}
\label{sec:classification}

Tab.~\ref{tab:classification_table} shows the accuracy on ModelNet40 and ScanObjectNN using 2,048-point inputs for all architectures compared. Classical deep learning methods achieve the strongest performance, with PointNet and PointMLP reaching 91.6\% on ModelNet40, whereas PointNet++ achieves 81.7\% on ScanObjectNN. Among transformer-based architectures, our method leads with 90.6\% on ModelNet40 and 80.0\% on ScanObjectNN (using our small variant), surpassing other transformer approaches, and converging visibly faster.

Our controlled training setup, without learning rate scheduling and hyperparameter tuning, may particularly impact transformer-based models, which typically benefit from longer training schedules and careful optimization strategies. This observation suggests that the superior performance of classical architectures in our experiments might partially originate from their robustness to simplified training protocols.

\subsubsection{Pre-training}
\label{sec:pretraining}

Next, we evaluate only transformer-based architectures on the Objaverse-LVIS subset, with results summarized in Table~\ref{tab:pretraining}. Despite the modest size of this curated dataset, our "small" model achieves an accuracy of 36.4\%, and closely matches PCT~\cite{guo2021pct} (36.3\%) in a medium-sized variant. As before, we observe that our approach typically converges faster than other transformer baselines.

\begin{figure}[t]
    \centering
    \begin{minipage}{0.37\textwidth}
        \includegraphics[width=\linewidth]{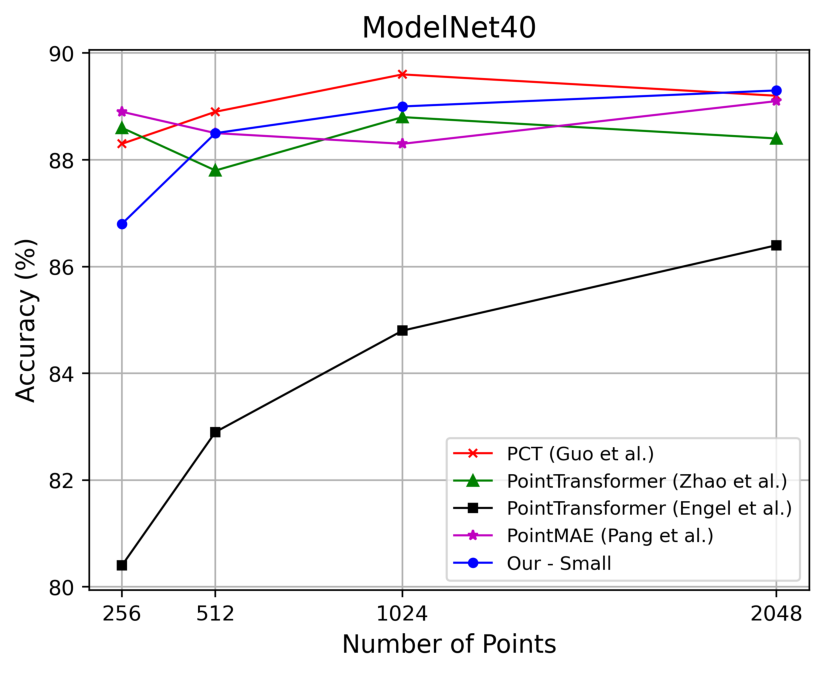}
    \end{minipage}%
    \begin{minipage}{0.58\textwidth}
        \caption{\textbf{Classification accuracy on ModelNet40 as a function of input point cloud size.} Models were trained for 30 epochs under identical conditions, with results showing the peak accuracy achieved. While PCT demonstrates superior performance in the 256-1024 point range, our architecture achieves competitive results and attains 89.3\% for 2048 points.
        }
    \label{fig:input_points}
    \end{minipage}
\end{figure}

\subsubsection{Zero-shot Evaluation}
\label{sec:zero_shot}

Table~\ref{tab:zero_shot} presents our zero-shot transfer evaluation, where we use the 30th epoch checkpoint from Objaverse-LVIS as a foundational model, tested directly on ModelNet40 and ScanObjectNN without fine-tuning. For this, we discard the classification head, extract final-layer features from transformer models pre-trained on the Objaverse-LVIS split (39k point clouds), and classify test samples by cosine similarity to class prototypes derived from the target dataset’s training set means. We report Top-1, Top-3, and Top-5 accuracies, benchmarking our reproduced transformers (e.g., PCT, PointTransformer variants, PointMAE, and our Small/Base models) against literature-reported foundation models. Strikingly, our method—trained solely on 39k shapes—outperforms all prior models pre-trained on ShapeNet’s 200k point clouds, even those with additional modalities or self-supervised objectives. Notably, Uni3D leverages a vast dataset of nearly one million 3D shapes, 10 million images, and 70 million texts, with shapes rendered from 10 viewpoints, yet our lightweight approach narrows the gap to state-of-the-art performance. Furthermore, the end-to-end training employed by our model, a characteristic shared with top-performing foundation models like Uni3D, may be a key contributor to these strong generalization capabilities. We also suspect this strong performance, despite the smaller dataset size, is partly attributable to the high quality and clean labeling of our pre-training data.

\subsubsection{Ablation Studies}
\label{sec:ablations}

As shown in Figure~\ref{fig:input_points}, our first ablation study examines how classification performance scales with different point cloud resolutions on ModelNet40. While PCT excels in the 256–1024 point range, our architecture remains competitive and achieves a peak accuracy of 89.3\% at 2048 points, slightly surpassing previous works. These findings highlight that model effectiveness can vary considerably with resolution, reflecting distinct scaling behaviors across architectures.

Our ablation (Table~\ref{tab:ablation}) shows simple token embedding summation outperforms a linear merging layer in the hierarchical transformer, suggesting additive merging better preserves geometric relationships without unnecessary parameters. Notably, the non-hierarchical transformer also achieves competitive results, likely due to the advantages of the chosen patching scheme and the robust positional embeddings that effectively guide the attention mechanism. Although we observed higher accuracy at 30 epochs with 32 and 128 patches, in practice, we chose 64 patches to balance computational efficiency and representational capacity. Furthermore, increasing embedding dimension and attention heads consistently improves performance.

\begin{table}[t!]
\caption{\textbf{Ablation Study of Pointy Design Choices.} We evaluate the impact of key architectural components—hierarchical transformer modules, token merging strategies (Addition vs. Linear), positional embeddings, and activation functions (GeLU vs. ReLU)—as well as model configuration parameters including embedding dimension, number of attention heads, patch count, and points per patch. For each variant, we report the best test accuracy (in \%) on ModelNet40 over 30 training epochs (training setup detailed in Section~\ref{sec:implementation}). All experiments are conducted with 2048 input points.}
\centering
\adjustbox{max width=\linewidth}{
\begin{tabular}{@{}ccccccccc@{}}
\toprule
\multicolumn{4}{c}{Architecture} & \multicolumn{4}{c}{Model Configuration} & \multicolumn{1}{l}{ModelNet40} \\
\cmidrule(r){1-4} \cmidrule(r){5-8} \cmidrule(l){9-9}
Hierarchical & Token Merging & Positional & Activation & Embedding & Attention & Patch & Points per & OA \\
Transformer & Strategy & Embedding & Function & Dimension & Heads & Count & Patch & (\%) $\uparrow$ \\
\midrule
\checkmark & Linear     & \checkmark & GeLU & 192 & 64 & 64  & 32 & 88.0 \\
\checkmark & Addition   & \checkmark & GeLU & 192 & 64 & 64  & 32 & 88.9 \\
\checkmark & Addition   & \checkmark & GeLU & 192 & 32 & 64  & 32 & 88.9 \\
\checkmark & Addition   & \checkmark & GeLU & 192 & 16 & 64  & 32 & 89.2 \\
\texttimes & \texttimes & \checkmark & GeLU & 192 & 64 & 64  & 32 & 89.3 \\
\checkmark & Addition   & \checkmark & ReLU & 192 & 64 & 128 & 32 & 89.5 \\
\checkmark & Addition   & \checkmark & GeLU & 192 & 64 & 64  & 32 & 89.5 \\
\checkmark & Addition   & \texttimes & GeLU & 192 & 64 & 64  & 32 & 89.6 \\
\checkmark & Addition   & \checkmark & GeLU & 510 & 64 & 64  & 32 & 89.7 \\
\checkmark & Addition   & \checkmark & GeLU & 192 & 64 & 128 & 32 & 89.8 \\
\checkmark & Addition   & \checkmark & GeLU & 192 & 64 & 32  & 64 & 90.0 \\
\bottomrule
\end{tabular}
}
\label{tab:ablation}
\end{table}

\section{Conclusion}
\label{sec:conclusion}

In this paper, we introduced a lightweight, transformer-based backbone for learning point cloud representations and conducted a comprehensive replication study under a unified training and benchmarking regime. Our experimental results demonstrate that even with only 39k pre-training samples, the proposed architecture achieves competitive and often superior performance compared to much larger foundation models trained on over 200k point clouds. Moreover, our model approaches the performance of state-of-the-art methods that leverage over one million point clouds, images, and text samples. These results underscore the importance of carefully selecting the training setup, architecture, end-to-end optimization, and data curation, rather than relying solely on extensive cross-modal supervision or massive datasets.
Finally, our results provide a transparent comparison of architectural choices and highlight the promise of simple yet powerful backbones. We hope that our open-source code, training protocols, and pre-trained weights will inspire broader engagement with controlled experimentation in 3D representation learning and ultimately promote reproducible and efficient approaches to foundation models in this rapidly evolving domain.

While the zero-shot results on classification tasks are encouraging, they also highlight an important limitation: our model (and all other reproductions) has been trained primarily under a classification objective. This may limit the model's ability to fully exploit the 3D geometry for finer-grained or denser prediction tasks. Furthermore, pre-training primarily on a curated, less noisy dataset might necessitate additional fine-tuning for optimal generalization to more challenging real-world scanner data. 

Therefore, our future research should investigate other settings, in particular semantic and instance segmentation in real 3D scenes, training on larger datasets, and verifying pre-training strategies and multimodal alignments (e.g., with images or text) in a similarly controlled experimental framework.

\subsubsection{Acknowledgments} 
This is a post-peer-review, pre-copyedit version of an article to be published in the proceedings of the \textit{Advanced Concepts for Intelligent Vision Systems (ACIVS 2025)}. 
An earlier version of this work was also presented at the \textit{Open Science for Foundation Models (SCI-FM)} workshop at ICLR 2025. 

\smallskip 
This work was supported by the National Science Centre, Poland, under research project no UMO-2023/51/B/ST6/01646. We would also like to thank Dominik Pieczyński from PUT Vision Lab for his assistance in managing and setting up the computational resources used in this study.

\bibliographystyle{plain}
\bibliography{iclr2025_conference}

@inproceedings{qi2017pointnet,
  title={Pointnet: Deep learning on point sets for 3d classification and segmentation},
  author={Qi, Charles R and Su, Hao and Mo, Kaichun and Guibas, Leonidas J},
  booktitle={Proceedings of the IEEE conference on computer vision and pattern recognition},
  pages={652--660},
  year={2017}
}

@article{qi2017pointnet++,
  title={Pointnet++: Deep hierarchical feature learning on point sets in a metric space},
  author={Qi, Charles Ruizhongtai and Yi, Li and Su, Hao and Guibas, Leonidas J},
  journal={Advances in neural information processing systems},
  volume={30},
  year={2017}
}

@article{wang2019dynamic,
  title={Dynamic graph cnn for learning on point clouds},
  author={Wang, Yue and Sun, Yongbin and Liu, Ziwei and Sarma, Sanjay E and Bronstein, Michael M and Solomon, Justin M},
  journal={ACM Transactions on Graphics (tog)},
  volume={38},
  number={5},
  pages={1--12},
  year={2019},
  publisher={Acm New York, NY, USA}
}

@inproceedings{ma2022rethinking,
title={Rethinking Network Design and Local Geometry in Point Cloud: A Simple Residual {MLP} Framework},
author={Xu Ma and Can Qin and Haoxuan You and Haoxi Ran and Yun Fu},
booktitle={International Conference on Learning Representations},
year={2022},
url={https://openreview.net/forum?id=3Pbra-\_u76D}
}

@article{guo2021pct,
  title={Pct: Point cloud transformer},
  author={Guo, Meng-Hao and Cai, Jun-Xiong and Liu, Zheng-Ning and Mu, Tai-Jiang and Martin, Ralph R and Hu, Shi-Min},
  journal={Computational Visual Media},
  volume={7},
  pages={187--199},
  year={2021},
  publisher={Springer}
}

@inproceedings{zhao2021point,
  title={Point transformer},
  author={Zhao, Hengshuang and Jiang, Li and Jia, Jiaya and Torr, Philip HS and Koltun, Vladlen},
  booktitle={Proceedings of the IEEE/CVF international conference on computer vision},
  pages={16259--16268},
  year={2021}
}

@article{engel2021point,
  title={Point transformer},
  author={Engel, Nico and Belagiannis, Vasileios and Dietmayer, Klaus},
  journal={IEEE access},
  volume={9},
  pages={134826--134840},
  year={2021},
  publisher={IEEE}
}

@inproceedings{pang2022masked,
  title={Masked autoencoders for point cloud self-supervised learning},
  author={Pang, Yatian and Wang, Wenxiao and Tay, Francis EH and Liu, Wei and Tian, Yonghong and Yuan, Li},
  booktitle={Computer Vision--ECCV 2022: 17th European Conference, Tel Aviv, Israel, October 23--27, 2022, Proceedings, Part II},
  pages={604--621},
  year={2022},
  organization={Springer}
}

@inproceedings{wu20153d,
  title={3d shapenets: A deep representation for volumetric shapes},
  author={Wu, Zhirong and Song, Shuran and Khosla, Aditya and Yu, Fisher and Zhang, Linguang and Tang, Xiaoou and Xiao, Jianxiong},
  booktitle={Proceedings of the IEEE conference on computer vision and pattern recognition},
  pages={1912--1920},
  year={2015}
}

@inproceedings{uy2019revisiting,
  title={Revisiting point cloud classification: A new benchmark dataset and classification model on real-world data},
  author={Uy, Mikaela Angelina and Pham, Quang-Hieu and Hua, Binh-Son and Nguyen, Thanh and Yeung, Sai-Kit},
  booktitle={Proceedings of the IEEE/CVF international conference on computer vision},
  pages={1588--1597},
  year={2019}
}

@inproceedings{deitke2023objaverse,
  title={Objaverse: A universe of annotated 3d objects},
  author={Deitke, Matt and Schwenk, Dustin and Salvador, Jordi and Weihs, Luca and Michel, Oscar and VanderBilt, Eli and Schmidt, Ludwig and Ehsani, Kiana and Kembhavi, Aniruddha and Farhadi, Ali},
  booktitle={Proceedings of the IEEE/CVF Conference on Computer Vision and Pattern Recognition},
  pages={13142--13153},
  year={2023}
}

@article{graham2017submanifold,
  title={3D Semantic Segmentation with Submanifold Sparse Convolutional Networks},
  author={Graham, Benjamin and Engelcke, Martin and van der Maaten, Laurens},
  journal={CVPR},
  year={2018}
}

@inproceedings{qi2023contrast,
  title={Contrast with reconstruct: Contrastive 3d representation learning guided by generative pretraining},
  author={Qi, Zekun and Dong, Runpei and Fan, Guofan and Ge, Zheng and Zhang, Xiangyu and Ma, Kaisheng and Yi, Li},
  booktitle={International Conference on Machine Learning},
  pages={28223--28243},
  year={2023},
  organization={PMLR}
}

@inproceedings{yu2022point,
  title={Point-bert: Pre-training 3d point cloud transformers with masked point modeling},
  author={Yu, Xumin and Tang, Lulu and Rao, Yongming and Huang, Tiejun and Zhou, Jie and Lu, Jiwen},
  booktitle={Proceedings of the IEEE/CVF conference on computer vision and pattern recognition},
  pages={19313--19322},
  year={2022}
}

@inproceedings{xue2024ulip,
  title={Ulip-2: Towards scalable multimodal pre-training for 3d understanding},
  author={Xue, Le and Yu, Ning and Zhang, Shu and Panagopoulou, Artemis and Li, Junnan and Mart{\'\i}n-Mart{\'\i}n, Roberto and Wu, Jiajun and Xiong, Caiming and Xu, Ran and Niebles, Juan Carlos and others},
  booktitle={Proceedings of the IEEE/CVF Conference on Computer Vision and Pattern Recognition},
  pages={27091--27101},
  year={2024}
}

@article{liu2024openshape,
  title={Openshape: Scaling up 3d shape representation towards open-world understanding},
  author={Liu, Minghua and Shi, Ruoxi and Kuang, Kaiming and Zhu, Yinhao and Li, Xuanlin and Han, Shizhong and Cai, Hong and Porikli, Fatih and Su, Hao},
  journal={Advances in neural information processing systems},
  volume={36},
  year={2024}
}

@inproceedings{maoopendlign,
  title={OpenDlign: Open-World Point Cloud Understanding with Depth-Aligned Images},
  author={Mao, Ye and Jing, Junpeng and Mikolajczyk, Krystian},
  booktitle={The Thirty-eighth Annual Conference on Neural Information Processing Systems},
  year={2024}
}

@InProceedings{Zhang_2024_CVPR,
    author    = {Zhang, Zhihao and Cao, Shengcao and Wang, Yu-Xiong},
    title     = {TAMM: TriAdapter Multi-Modal Learning for 3D Shape Understanding},
    booktitle = {Proceedings of the IEEE/CVF Conference on Computer Vision and Pattern Recognition (CVPR)},
    month     = {June},
    year      = {2024},
    pages     = {21413-21423}
}

@inproceedings{zhou2023uni3d,
  title={Uni3d: Exploring unified 3d representation at scale},
  author={Zhou, Junsheng and Wang, Jinsheng and Ma, Baorui and Liu, Yu-Shen and Huang, Tiejun and Wang, Xinlong},
  booktitle={International Conference on Learning Representations (ICLR)},
  year={2024}
}

@inproceedings{he2015delving,
  title={Delving deep into rectifiers: Surpassing human-level performance on imagenet classification},
  author={He, Kaiming and Zhang, Xiangyu and Ren, Shaoqing and Sun, Jian},
  booktitle={Proceedings of the IEEE international conference on computer vision},
  pages={1026--1034},
  year={2015}
}

@article{vaswani2017attention,
  title={Attention is all you need},
  author={Vaswani, A},
  journal={Advances in Neural Information Processing Systems},
  year={2017}
}

@article{guo2023point,
  title={Point-bind \& point-llm: Aligning point cloud with multi-modality for 3d understanding, generation, and instruction following},
  author={Guo, Ziyu and Zhang, Renrui and Zhu, Xiangyang and Tang, Yiwen and Ma, Xianzheng and Han, Jiaming and Chen, Kexin and Gao, Peng and Li, Xianzhi and Li, Hongsheng and others},
  journal={arXiv preprint arXiv:2309.00615},
  year={2023}
}

@inproceedings{xu2024pointllm,
  title={Pointllm: Empowering large language models to understand point clouds},
  author={Xu, Runsen and Wang, Xiaolong and Wang, Tai and Chen, Yilun and Pang, Jiangmiao and Lin, Dahua},
  booktitle={European Conference on Computer Vision},
  pages={131--147},
  year={2024},
  organization={Springer}
}

@article{liu2024uni3d,
  title={Uni3d-llm: Unifying point cloud perception, generation and editing with large language models},
  author={Liu, Dingning and Huang, Xiaoshui and Hou, Yuenan and Wang, Zhihui and Yin, Zhenfei and Gong, Yongshun and Gao, Peng and Ouyang, Wanli},
  journal={arXiv preprint arXiv:2402.03327},
  year={2024}
}

@inproceedings{liu2021swin,
  title={Swin transformer: Hierarchical vision transformer using shifted windows},
  author={Liu, Ze and Lin, Yutong and Cao, Yue and Hu, Han and Wei, Yixuan and Zhang, Zheng and Lin, Stephen and Guo, Baining},
  booktitle={Proceedings of the IEEE/CVF international conference on computer vision},
  pages={10012--10022},
  year={2021}
}

@article{oquab2024dinov2,
title={{DINO}v2: Learning Robust Visual Features without Supervision},
author={Maxime Oquab and Timoth{\'e}e Darcet and Th{\'e}o Moutakanni and Huy V. Vo and Marc Szafraniec et al.},
journal={Transactions on Machine Learning Research},
issn={2835-8856},
year={2024},
url={https://openreview.net/forum?id=a68SUt6zFt},
note={Featured Certification}
}

@article{devlin2018bert,
  title={Bert: Pre-training of deep bidirectional transformers for language understanding},
  author={Devlin, Jacob},
  journal={arXiv preprint arXiv:1810.04805},
  year={2018}
}

@article{radford2019language,
  title={Language models are unsupervised multitask learners},
  author={Radford, Alec and Wu, Jeffrey and Child, Rewon and Luan, David and Amodei, Dario and Sutskever, Ilya and others},
  journal={OpenAI blog},
  volume={1},
  number={8},
  pages={9},
  year={2019}
}

@article{brown2020language,
  title={Language models are few-shot learners},
  author={Brown, Tom and Mann, Benjamin and Ryder, Nick and Subbiah, Melanie and Kaplan, Jared D and Dhariwal, Prafulla and Neelakantan, Arvind and Shyam, Pranav and Sastry, Girish and Askell, Amanda and others},
  journal={Advances in neural information processing systems},
  volume={33},
  pages={1877--1901},
  year={2020}
}

@inproceedings{he2016deep,
  title={Deep residual learning for image recognition},
  author={He, Kaiming and Zhang, Xiangyu and Ren, Shaoqing and Sun, Jian},
  booktitle={Proceedings of the IEEE conference on computer vision and pattern recognition},
  pages={770--778},
  year={2016}
}

@inproceedings{DosovitskiyB0WZ21,
  author       = {Alexey Dosovitskiy and
                  Lucas Beyer and
                  Alexander Kolesnikov and
                  Dirk Weissenborn and
                  Xiaohua Zhai and
                  Thomas Unterthiner and
                  Mostafa Dehghani and
                  Matthias Minderer and
                  Georg Heigold and
                  Sylvain Gelly and
                  Jakob Uszkoreit and
                  Neil Houlsby},
  title        = {An Image is Worth 16x16 Words: Transformers for Image Recognition
                  at Scale},
  booktitle    = {9th International Conference on Learning Representations, {ICLR} 2021,
                  Virtual Event, Austria, May 3-7, 2021},
  publisher    = {OpenReview.net},
  year         = {2021},
  url          = {https://openreview.net/forum?id=YicbFdNTTy},
  bibsource    = {dblp computer science bibliography, https://dblp.org}
}

@inproceedings{goswami2024moment,
  title={MOMENT: A Family of Open Time-series Foundation Models},
  author={Mononito Goswami and Konrad Szafer and Arjun Choudhry and Yifu Cai and Shuo Li and Artur Dubrawski},
  booktitle={International Conference on Machine Learning},
  year={2024}
}

\end{document}